\documentclass[12pt]{article}

\usepackage{microtype}
\usepackage{float}
\usepackage[utf8]{inputenc}
\usepackage{amsmath}
\usepackage{graphicx}
\usepackage{booktabs}
\usepackage{hyperref}
\usepackage{caption}
\usepackage{float}
\usepackage{geometry}
\usepackage{authblk}

\title{The Relationship Between Head Injury and Alzheimer’s Disease: A Causal Analysis with Bayesian Networks}

\author[1]{Andrei Lixandru}
\affil[1]{Radboud University, Department of Machine Learning and Neural Computing, The Netherlands}

\date{\today}

\begin{document}

\maketitle

\begin{abstract}
This study examines the potential causal relationship between head injury and the risk of developing Alzheimer's disease (AD) using Bayesian networks and regression models. Using a dataset of 2,149 patients, we analyze key medical history variables, including head injury history, memory complaints, cardiovascular disease, and diabetes. Logistic regression results suggest an odds ratio of 0.88 for head injury, indicating a potential but statistically insignificant protective effect against AD. In contrast, memory complaints exhibit a strong association with AD, with an odds ratio of 4.59. Linear regression analysis further confirms the lack of statistical significance for head injury (coefficient: -0.0245, p = 0.469) while reinforcing the predictive importance of memory complaints. These findings highlight the complex interplay of medical history factors in AD risk assessment and underscore the need for further research utilizing larger datasets and advanced causal modeling techniques.
\end{abstract}

\tableofcontents

\section{Introduction}

Traumatic brain injury (TBI) has been strongly associated with an elevated risk of developing Alzheimer's disease (AD), with studies reporting that moderate and severe TBIs increase this risk by factors of 2.3 and 4.5, respectively \cite{gottlieb2000head}. Understanding the causal relationship between head injuries and AD is critical for devising preventive strategies and interventions. This study investigates the causal effect of head injuries on the likelihood of developing AD, incorporating multiple factors from patients' medical histories. The analysis utilizes Bayesian networks and Directed Acyclic Graphs (DAGs), which are well-suited to model causal relationships under uncertainty.

The primary research question guiding this study is: \textit{"When considering multiple medical history factors, what is the causal effect of head injury on the risk of developing Alzheimer's disease?"} To address this question, the study employs a DAG-based approach for causal inference, leveraging a detailed health dataset to explore the relationships between a history of head injury (the exposure variable) and AD diagnosis (the outcome variable).

This research builds upon prior work outlined in Assignment 1, which established the foundational DAG and provided the theoretical framework for the causal analysis. The DAG incorporates key variables such as family history of AD, depression, cardiovascular disease, diabetes, hypertension, and memory complaints, supported by existing evidence. For instance, head injury is causally linked to depression \cite{jorge2004major}, cardiovascular disease \cite{stewart2022association}, and memory complaints \cite{izzy2023long}. Additionally, depression has established associations with cardiovascular disease \cite{khandaker2020shared} and hypertension \cite{inoue2024depressive}. The dataset and methods are carefully chosen to ensure rigorous testing of these causal hypotheses.

By investigating these complex relationships, this study aims to provide new insights into the role of head injuries in the development of Alzheimer's disease, while contributing to the broader understanding of risk factors for AD. The subsequent subsections detail the dataset, Bayesian network design, and validation methods employed to assess the causal structure. 

\subsection{Dataset Overview} We use a Kaggle-based dataset from El Kharoua et al. \cite{rabie_el_kharoua_2024}, containing health data for 2,149 patients, including demographics, medical history, and clinical measurements. 

We chose this dataset because this was the largest AD dataset that included a history of head injury as a variable. We only include binary variables because the focus of the study were variables related to the medical history of the patient, which happened to be only binary variables.

\begin{table}[H]
    \centering
    \caption{Dataset Variable Description}
    \begin{tabular}{|l|l|c|c|}
        \hline
        \textbf{Variable Name} & \textbf{Variable Type} & \textbf{Number of Levels} \\
        \hline
        FamilyHistoryAlzheimers & Categorical & 2 (0, 1)   \\
        Depression              & Categorical & 2 (0, 1)   \\
        HeadInjury              & Categorical & 2 (0, 1)   \\
        CardiovascularDisease              & Categorical & 2 (0, 1) \\
        Diabetes              & Categorical & 2 (0, 1) \\
        Hypertension              & Categorical & 2 (0, 1)  \\
        MemoryComplaints              & Categorical & 2 (0, 1) \\

        Diagnosis & Categorical & 2 (0, 1) \\
        \hline
    \end{tabular}
    \label{tab:variable_description}
\end{table}

\subsection{Bayesian Network and Model Design} 
\begin{figure}[H]
    \centering
    \includegraphics[width=0.8\textwidth]{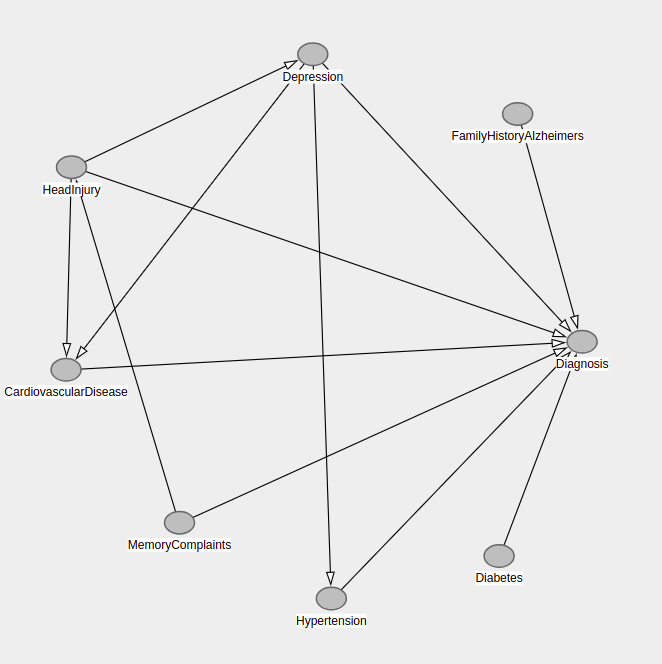} 
    \caption{First proposed DAG for our problem.}
    \label{fig:first_DAG} 
\end{figure}

The Bayesian network focuses on medical history, as head injury is central to this analysis. Each variable was modeled as a predictor of AD diagnosis, with additional dependencies derived from domain knowledge. Evidence supports causal links between head injury and cardiovascular disease \cite{stewart2022association}, depression \cite{jorge2004major}, and memory complaints \cite{izzy2023long}, as well as links between depression and cardiovascular disease \cite{khandaker2020shared} and hypertension \cite{inoue2024depressive}. The initial DAG (Figure~\ref{fig:first_DAG}) reflects these relationships.

\subsection{Validation of Network Structure} To test the independence assumptions in the DAG, we performed Chi-squared independence tests for FamilyHistoryAlzheimers, HeadInjury, and Diabetes. High p-values (e.g., 0.31, 0.42, and 1) confirmed no significant associations between these variables. This supports the DAG's accuracy in representing the relationships among predictors, requiring no restructuring.

\section{Methods}

\subsection{Logistic Regression for Odds Ratios}

Logistic regression is employed to estimate the causal relationship between head injury and Alzheimer's disease (AD) diagnosis, accounting for potential confounding factors. This method models the log-odds of the binary outcome variable (e.g., AD diagnosis) as a linear combination of predictor variables, allowing for the estimation of odds ratios (ORs) that quantify the strength and direction of associations.

The general logistic regression model used in this study is expressed as:

\begin{equation}
    \log \left( \frac{P(Y=1 \mid X, C)}{1 - P(Y=1 \mid X, C)} \right) = \beta_0 + \beta_1 X + \beta_2 C
\end{equation}

where:
\( P(Y=1 \mid X, C) \) represents the probability of being diagnosed with AD.
\( X \) denotes the predictor variable of interest, \textbf{HeadInjury}.
\( C \) represents the confounding variable, \textbf{MemoryComplaints}.
\( \beta_1 \) and \( \beta_2 \) are the regression coefficients for \textbf{HeadInjury} and \textbf{MemoryComplaints}, respectively.

The odds ratio (OR) for the predictor variable \( X \) is calculated as:

\[
OR_X = \exp(\beta_1)
\]

This OR quantifies the multiplicative change in the odds of AD diagnosis for a one-unit increase in \( X \), adjusting for \( C \). An OR greater than 1 indicates a positive association, while an OR less than 1 indicates a negative association.

To account for potential confounding, \textbf{MemoryComplaints} (\( C \)) was included in the model. By adjusting for this variable, the analysis estimates the adjusted odds ratio for \textbf{HeadInjury}, providing insights into its direct effect on AD diagnosis while controlling for the influence of memory complaints.

The statistical significance of the odds ratios was assessed using 95\% confidence intervals (CI). If the CI for an OR excludes 1, the result is considered statistically significant at the 5\% level. The logistic regression analysis was conducted using the dataset described earlier.

In this study, the odds ratios for \textbf{HeadInjury} and \textbf{MemoryComplaints} were calculated to explore their respective associations with the likelihood of an AD diagnosis. The results and their implications are detailed in the Results section.

\subsection{Linear Regression for Risk Difference}

Linear regression was employed to estimate the causal effect of head injury on the risk of being diagnosed with Alzheimer's disease (AD), with the \textit{risk difference} (RD) serving as the measure of effect. The RD quantifies the difference in the probability of the outcome between exposed and unexposed groups. This analysis adjusts for potential confounders to provide an unbiased estimate of the causal relationship between head injury and Alzheimer's diagnosis.

The relationship between the binary outcome \( Y \) (e.g., Alzheimer's diagnosis) and the binary exposure \( X \) (e.g., head injury) was modeled using the following linear regression equation:

\[
Y = \beta_0 + \beta_1 X + \sum_{i=1}^{k} \beta_{i+1} Z_i + \epsilon
\]

where:
\( Y \) is the binary outcome variable, representing Alzheimer's diagnosis (\( Y = 1 \) for diagnosed, \( Y = 0 \) for not diagnosed),
\( X \) is the binary exposure variable, representing a history of head injury (\( X = 1 \) for exposed, \( X = 0 \) for unexposed),
\( Z_1, Z_2, \dots, Z_k \) are confounder variables (e.g., memory complaints),
\( \beta_0 \) is the intercept term,
\( \beta_1 \) is the coefficient for the exposure variable, representing the causal effect of head injury on Alzheimer's diagnosis,
\( \beta_{i+1} \) are the coefficients for the confounder variables, and
\( \epsilon \) is the error term, accounting for unobserved factors influencing \( Y \).

The coefficient \( \hat{\beta}_1 \) represents the risk difference (RD), which is the difference in the probability of the outcome between exposed and unexposed groups, adjusted for confounders. Specifically, the risk for the exposed group (\( X = 1 \)) is calculated as:

\[
\text{Risk}_{X=1} = P(Y = 1 \mid X = 1) = \beta_0 + \beta_1
\]

The risk for the unexposed group (\( X = 0 \)) is:

\[
\text{Risk}_{X=0} = P(Y = 1 \mid X = 0) = \beta_0
\]

The risk difference is then given by:

\[
\text{RD} = \text{Risk}_{X=1} - \text{Risk}_{X=0} = (\beta_0 + \beta_1) - \beta_0 = \beta_1
\]

This approach assumes a linear relationship between the predictors and the outcome, which enables \( \beta_1 \) to directly represent the adjusted RD. Adjusting for confounders \( Z_1, Z_2, \dots, Z_k \) is essential to control for biases arising from variables that influence both the exposure \( X \) and the outcome \( Y \).

In this analysis, the inclusion of confounders such as memory complaints ensured that the estimated \( \beta_1 \) reflects the causal effect of head injury on the risk of Alzheimer's diagnosis. The RD provides an interpretable estimate of the change in the probability of the outcome due to the exposure, adjusted for confounders. A positive RD indicates an increased probability of Alzheimer's diagnosis in the exposed group, while a negative RD suggests a decreased probability.

\section{Results}
\subsection{Logistic Regression for Odds Ratios}

The logistic regression analysis was conducted to estimate the causal effect of head injury on the diagnosis of Alzheimer's disease (AD), adjusting for the potential confounding effect of memory complaints. The analysis produced an odds ratio (OR) for \textbf{HeadInjury} of 0.88, which suggests that having a head injury is associated with lower odds of receiving a diagnosis of Alzheimer's disease, compared to individuals without a head injury. Specifically, the odds of the outcome (AD diagnosis) decrease by 12\% for individuals with a head injury, as calculated by:

\[
1 - 0.88 = 0.12 \quad \text{or} \quad 12\%
\]

This indicates that the presence of a head injury may be a protective factor against the development of Alzheimer's disease, although this relationship should be interpreted with caution due to the following considerations.

The confidence interval (CI) for the odds ratio of \textbf{HeadInjury} ranges from 0.63 to 1.22. Since this interval includes the value 1, the result is not statistically significant at the 5\% level. In other words, while the point estimate suggests a protective effect, the wide CI indicates that the true effect could range from a substantial protective effect to no effect at all. Thus, we cannot confidently conclude that head injury has a causal effect on the likelihood of being diagnosed with Alzheimer's disease, given the statistical insignificance of the result.

In contrast, the odds ratio for \textbf{MemoryComplaints}, which was included as a confounder in the model, is 4.59 with a 95\% confidence interval of (3.69, 5.72). This suggests a statistically significant positive association between memory complaints and the likelihood of being diagnosed with Alzheimer's disease. Specifically, the ratio of odds of developing AD in the group of people with memory complaints versus developing AD in the group of people without memory complaints is 4.59. The confidence interval does not include 1, indicating that the effect is statistically significant at the 5\% level. Therefore, memory complaints are a significant predictor of Alzheimer's diagnosis in this model.

In conclusion, while the analysis suggests a negative (protective) association between head injury and Alzheimer's diagnosis, the lack of statistical significance prevents a strong causal claim. In contrast, memory complaints are clearly associated with an increased likelihood of diagnosis, and the evidence supports a significant causal effect of memory complaints on the diagnosis of Alzheimer's disease.

\begin{figure}[H]
    \centering
    \includegraphics[width=0.8\textwidth]{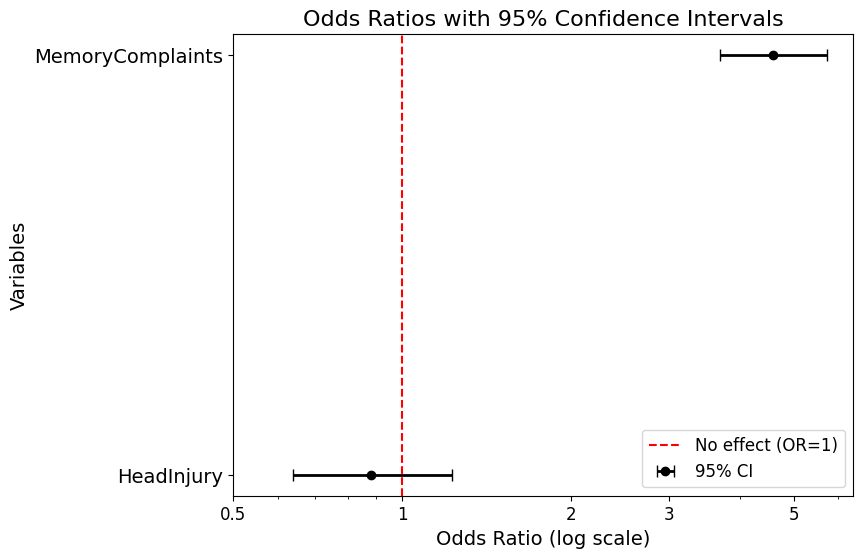}
    \caption{Odds Ratios with 95\% Confidence Intervals}
    \label{fig:odds_ratios_plot}
\end{figure}

A bar plot visualizes the odds ratios for \textbf{HeadInjury} and \textbf{MemoryComplaints} along with their 95\% confidence intervals, highlighting the strength and direction of their association with the likelihood of \textbf{Diagnosis}, with a reference line at OR=1 indicating no effect (see Figure~\ref{fig:odds_ratios_plot}).

\subsection{Linear Regression for Risk Difference}

The results of the linear regression analysis, which aimed to investigate the causal relationship between \textbf{HeadInjury} and the diagnosis of Alzheimer's disease, while accounting for the confounding effect of \textbf{MemoryComplaints}, are presented below. This analysis was grounded in the theoretical framework of linear regression for risk difference estimation, as outlined previously.
The estimated coefficient for \textbf{HeadInjury} in the presence of \textbf{MemoryComplaints} as a confounder was \(-0.0245\), with a corresponding \textit{p-value} of \(0.469\). This \textit{p-value} exceeds the conventional significance threshold of \(0.05\), which suggests that we fail to reject the null hypothesis. In statistical terms, this indicates that there is no statistically significant evidence to suggest a causal relationship between \textbf{HeadInjury} and the likelihood of receiving a diagnosis of Alzheimer's disease in this sample.

The negative coefficient of \(-0.0245\) indicates a very small, inverse relationship between \textbf{HeadInjury} and the diagnosis of Alzheimer's disease. However, due to the lack of statistical significance, we interpret this coefficient with caution. The observed negative effect may be due to random variation, rather than indicating a true causal association. Therefore, while the direction of the relationship suggests that \textbf{HeadInjury} might reduce the likelihood of Alzheimer's diagnosis, the statistical insignificance prevents us from making any firm conclusions about the presence of a causal link.
In contrast to \textbf{HeadInjury}, the coefficient for \textbf{MemoryComplaints} was \(0.36\), with a \textit{p-value} significantly less than \(0.05\). This result indicates that \textbf{MemoryComplaints} is a statistically significant predictor of the likelihood of receiving an Alzheimer's diagnosis, even after controlling for the potential confounding effect of \textbf{HeadInjury}. The positive coefficient of \(0.36\) suggests that individuals who report memory complaints are more likely to be diagnosed with Alzheimer's disease compared to those who do not report such complaints, holding the effect of \textbf{HeadInjury} constant.

The positive coefficient can be interpreted as follows: for each unit increase in \textbf{MemoryComplaints} (i.e., the presence of memory complaints), the probability of being diagnosed with Alzheimer's disease increases by approximately 36\%, relative to individuals without memory complaints. This result is consistent with the existing literature, which suggests that memory problems are often associated with the onset of Alzheimer's disease and may serve as a key predictor of the disease's progression.

Given the findings, we conclude that \textbf{MemoryComplaints} plays a meaningful role in explaining the variation in the likelihood of receiving an Alzheimer's diagnosis, while \textbf{HeadInjury} does not exhibit a statistically significant relationship with the outcome variable in this sample. The theoretical background on linear regression for risk difference estimation supports this interpretation, as the regression model accounts for confounders and provides an adjusted estimate of the causal effect.

Overall, the present analysis suggests that while \textbf{HeadInjury} might not be a significant predictor of Alzheimer's diagnosis in this context, \textbf{MemoryComplaints} is a strong and significant predictor. Future research could further explore the potential mechanisms underlying the relationship between memory complaints and Alzheimer's disease, while continuing to investigate the role of \textbf{HeadInjury} with larger samples or alternative modeling approaches.

\begin{figure}[H]
    \centering
    \includegraphics[width=0.8\textwidth]{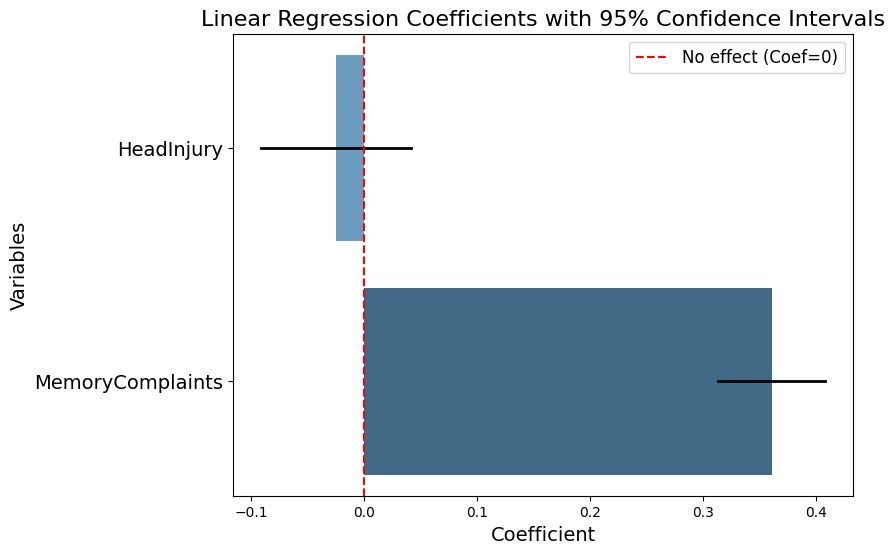}
    \caption{Linear Regression Coefficients with 95\% Confidence Intervals}
    \label{fig:coef_plot}
\end{figure}

Figure \ref{fig:coef_plot} illustrates the linear regression coefficients for \textbf{HeadInjury} and \textbf{MemoryComplaints}, along with their 95\% confidence intervals. The plot shows the magnitude and direction of the estimated effects of these variables on the likelihood of receiving a diagnosis of Alzheimer's disease, with the red vertical line at zero serving as a reference for no effect (coefficient = 0). The error bars represent the confidence intervals, highlighting the precision of the coefficient estimates.

\begin{figure}[H]
    \centering
    \includegraphics[width=0.8\textwidth]{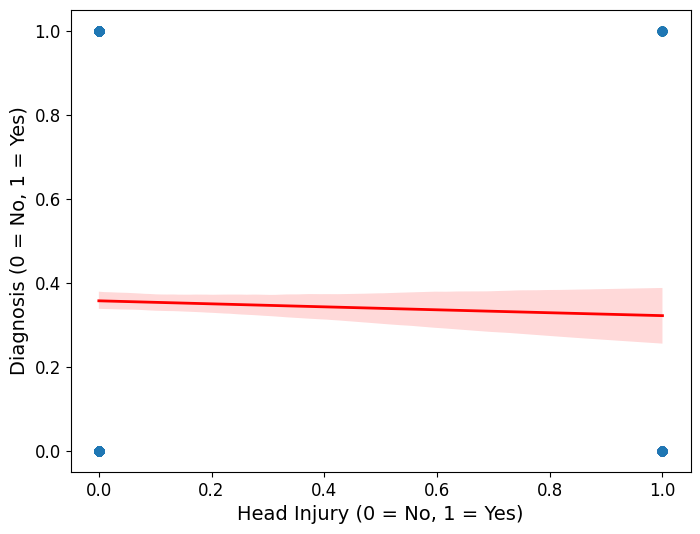}
    \caption{Regression plot of HeadInjury on Diagnosis (with MemoryComplaints control)}
    \label{fig:regression_plot}
\end{figure}

Figure \ref{fig:regression_plot} depicts the regression relationship between \textbf{HeadInjury} and \textbf{Diagnosis}, while controlling for the confounding effect of \textbf{MemoryComplaints}. The plot shows the predicted probability of receiving an Alzheimer's diagnosis based on the presence of head injury, with the red line representing the fitted logistic regression curve and the shaded red area the 95\% confidence interval. The x-axis indicates the presence (1) or absence (0) of head injury, and the y-axis represents the binary diagnosis outcome (1 for diagnosis, 0 for no diagnosis).

\section{Conclusion}

This study aimed to investigate the causal relationship between head injury and the risk of developing Alzheimer's disease (AD), utilizing both logistic and linear regression models. The logistic regression analysis indicated a protective effect of head injury on AD diagnosis with an odds ratio of 0.88. However, the 95\% confidence interval (0.63 to 1.22) included 1, leading to statistical insignificance. Similarly, the linear regression analysis for risk difference showed a small negative coefficient for head injury (\(-0.0245\)), but the p-value of 0.469 indicated no statistical significance.

In contrast, memory complaints were consistently found to be a strong predictor of AD diagnosis. The odds ratio for memory complaints was 4.59, and the regression coefficient was 0.36, both statistically significant. These results highlight the robust association between memory complaints and the likelihood of AD diagnosis.

While the analyses suggest a potential protective effect of head injury on AD, the findings are not statistically significant, and further research is needed to clarify this relationship. Memory complaints, however, emerge as a clear risk factor for Alzheimer's, emphasizing the need for further exploration of this variable in future studies.

\newpage
\label{sec:references}
\bibliographystyle{plain}
\bibliography{refs}

\newpage
\appendix
\section{Appendix}
\label{sec:appendix}
 Figure 1 was constructed using DAGitty \cite{daggity}. The source code used for all analyses can be found at \url{https://github.com/AndreiLix/Causal_Inference}. The source code contains data preprocessing for having only the variables we need for our analyses, Chi-squared tests for testing independence between variables, the code used for the analysis with the logistic and linear regression models and for generating the analysis plots.
\end{document}